\pdfoutput=1

\documentclass[11pt]{article}

\usepackage {acl}

\usepackage{times}
\usepackage{latexsym}

\usepackage[T1]{fontenc}

\usepackage[utf8]{inputenc}

\usepackage{microtype}

\usepackage{inconsolata}

\usepackage{graphicx}
\usepackage{url}
\usepackage{stfloats}
\usepackage{bm}
\usepackage{amsmath}
\usepackage{enumerate}
\usepackage{caption}
\usepackage{tabularx}
\usepackage{booktabs}
\usepackage{xcolor}
\usepackage{paralist}
\usepackage{multirow} 
\usepackage{enumitem}

\title{JUREX-4E: Juridical Expert-Annotated Four-Element Knowledge Base for Legal Reasoning}

\author{\bf
Huanghai Liu\textsuperscript{\rm 1}\thanks{These authors contributed equally to this work.},
Quzhe Huang\textsuperscript{\rm 2*},
Qingjing Chen\textsuperscript{\rm 3},
Yiran Hu\textsuperscript{\rm 1},\\
\bf Jiayu Ma\textsuperscript{\rm1},
Yun Liu\textsuperscript{\rm 1},
Weixing Shen\textsuperscript{\rm 1}\thanks{Corresponding Author.},
Yansong Feng\textsuperscript{\rm 2$\dagger$}\\
\textsuperscript{\rm 1}School of Law, Tsinghua University \\
\textsuperscript{\rm 2}Wangxuan Institute of Computer Technology, Peking University \\
\textsuperscript{\rm 3}Department of Legal Studies, University of Bologna\\
\texttt{\{liuhh23,chenqj21,huyr21,ma-jy24\}@mails.tsinghua.edu.cn}\\
\texttt{\{huangquzhe,fengyansong\}@pku.edu.cn}
\texttt{\{liuyun89,wxshen\}@mail.tsinghua.edu.cn}
}

\begin{document}
\maketitle

\begin{abstract}

In recent years, Large Language Models (LLMs) have been widely applied to legal tasks. To enhance their understanding of legal texts and improve reasoning accuracy, a promising approach is to incorporate legal theories. One of the most widely adopted theories is the Four-Element Theory (FET), which defines the crime constitution through four elements: Subject, Object, Subjective Aspect, and Objective Aspect. While recent work has explored prompting LLMs to follow FET, our evaluation demonstrates that LLM-generated four-elements are often incomplete and less representative, limiting their effectiveness in legal reasoning.
To address these issues, we present JUREX-4E, an expert-annotated four-element knowledge base covering 155 criminal charges. The annotations follow a progressive hierarchical framework grounded in legal source validity and incorporate diverse interpretive methods to ensure precision and authority. 
We evaluate JUREX-4E on the Similar Charge Disambiguation task and apply it to Legal Case Retrieval. Experimental results validate the high quality of JUREX-4E and its substantial impact on downstream legal tasks, underscoring its potential for advancing legal AI applications. The dataset and code are available at: \url{https://github.com/THUlawtech/JUREX}

\end{abstract}

\section{Introduction}

Large Language Models (LLMs) have recently demonstrated impressive performance in legal tasks such as charge prediction~\cite{yuan2024can} and legal case retrieval~\cite{feng2024legal}. In these applications, a key challenge is accurately understanding complex legal language.
To address this, recent studies have introduced legal theories into LLM workflows~\cite{jiang2023legal, servantez2024chain, yuan2024can, deng2023syllogistic}, as these theories provide structured reasoning frameworks and domain knowledge. Among these theories, the Four-Element Theory (FET) in Chinese criminal law~\cite{liang2017vicissitudes} is particularly important, as it defines the legal criteria for establishing criminal liability. FET breaks down a criminal charge into four elements: Subject, Object, Subjective Aspect, and Objective Aspect, which serve as the essential criteria for determining whether a defendant's behavior constitutes a specific crime.

Most current approaches rely on the LLM's internal knowledge to incorporate the FET. A common method is to ask LLMs to emulate expert reasoning processes. For example, designing four separate prompts to guide the LLM outputs in the form of four-elements~\cite{deng2023syllogistic}. This raises a critical question: Can LLMs reliably understand and apply the FET?

\begin{figure}[tp]
    \centering
    \includegraphics[width=1\columnwidth]{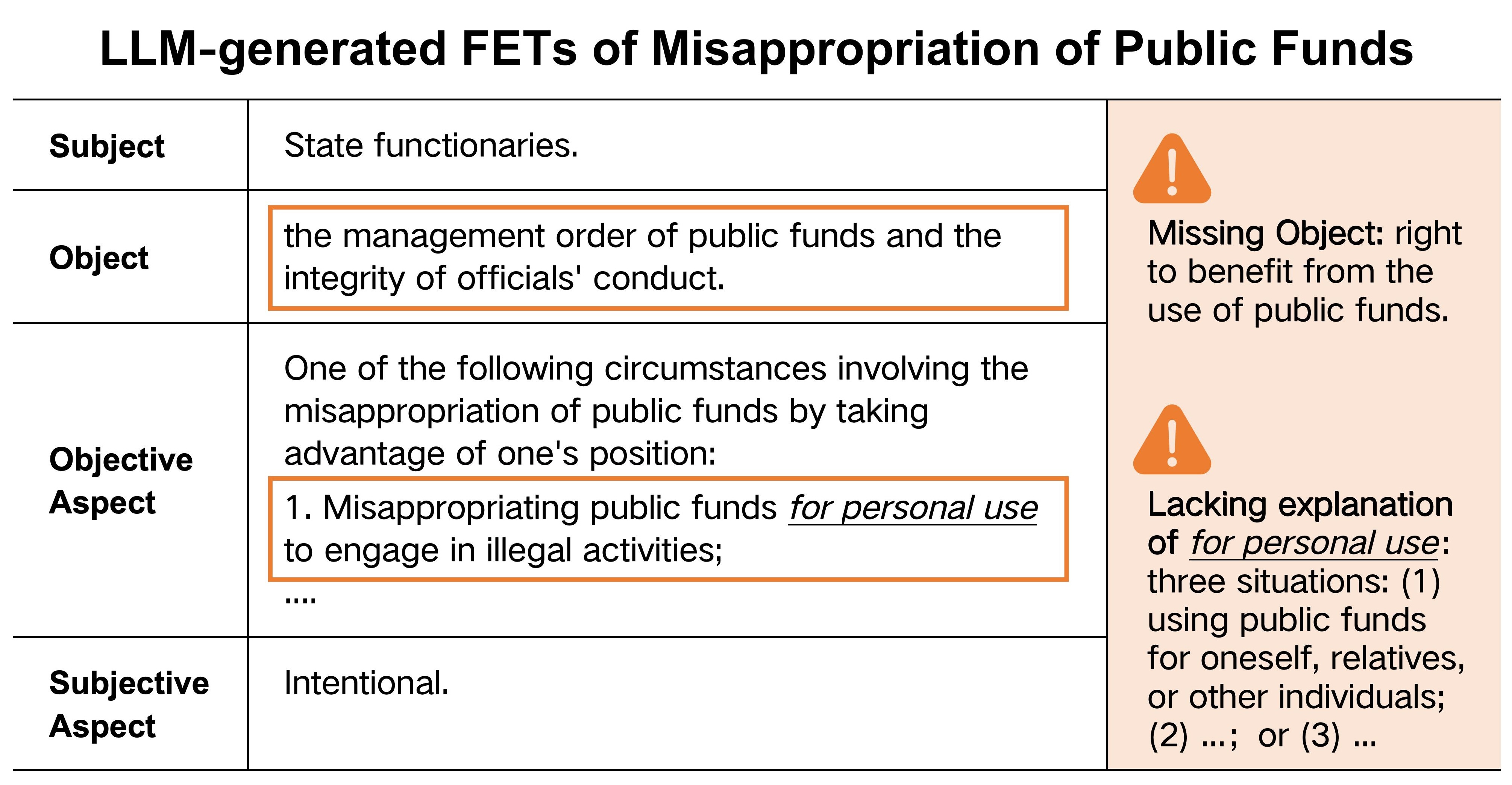}
    \caption{An example of LLM-generated four-elements.}
    \label{fig:example}
        \vspace{-1ex}
\end{figure}

To investigate this, we conducted a pilot study where we provide LLMs with legal articles and asked them to generate the four-elements for several representative charges~\cite{ouyang1999confusing}. Results show that the LLM-generated four-elements are often not accurate enough. As shown in Figure~\ref{fig:example}, in the charge of \textit{misappropriation of public funds}, the LLM failed to identify the right to benefit from the use of public funds, a core part of the Object. These results suggest that LLMs lack the domain knowledge and legal reasoning precision required for reliable FET application.

To help LLMs better utilize the FET in legal tasks, we construct \textbf{JUREX-4E: JUR}idical \textbf{EX}pert-annotated  \textbf{4-E}lement knowledge base for legal reasoning. The knowledge base covers 155 high-frequency criminal charges, each decomposed into Subject, Object, Subjective Aspect, and Objective Aspect. JUREX-4E is built through a four-stage Hierarchical Legal Interpretation System. In this process, legal experts refine each element by referencing sources in descending order of legal validity—Criminal Articles, Judicial Interpretations, Guiding Cases, and Academic Discourses—while applying appropriate interpretive methods at each stage. Each charge was annotated over a seven-month period, yielding knowledge-rich representations with an average annotation length of 472.5 words.


To assess the quality of JUREX-4E, we conduct a human evaluation on four independent dimensions, Precision, Completeness, Representativeness, and Standardization, grounded in legal scholarship on how criminal elements should be normatively defined and expressed in judicial contexts ~\cite{zhang2007normative}. The expert-annotated four-elements achieved an average score of 4.60 on a 5-point scale, significantly outperforming the LLM-generated ones, which scored 3.96. Among the four dimensions, the largest performance gaps appeared in Completeness and Representativeness, as expert annotations provided more comprehensive legal interpretations and summarized typical application scenarios, which are often overlooked by LLMs. 

To further evaluate the quality and utility of JUREX-4E, we conducted two downstream tasks: Similar Charge Disambiguation (SCD) and Legal Case Retrieval (LCR). In the SCD task~\cite{liu2021everything}, we tested whether different charges could be more effectively distinguished by incorporating four-element knowledge. Results show that expert-annotated four-elements from JUREX-4E consistently outperformed LLM-generated counterparts across various prompting strategies and model types, improving average accuracy by 0.70\% and F1-score by 0.75\%. In the LCR task~\cite{li2024lecardv2}, we incorporated JUREX-4E into the retrieval pipeline to guide case-level four-element generation and similarity matching, achieving better retrieval accuracy. Together, these findings validate the high quality and practical value of JUREX-4E in enhancing legal understanding and decision-making.

Our contributions are as follows:
\begin{compactenum}[(1)]
\item We demonstrate that while LLMs can assist legal reasoning to some extent, they still fall short in accurately understanding and applying the Four-Element Theory.

\item We construct the JUREX-4E, the first expert-annotated legal knowledge base grounded in a hierarchical legal interpretation framework based on legal source validity.

\item  We validate the quality and effectiveness of JUREX-4E on two representative legal tasks, Similar Charge Disambiguation (SCD) and Legal Case Retrieval (LCR), where it consistently outperforms LLM-generated representations across various prompting strategies.
\end{compactenum}

\section{Background}
The Four-Element Theory (FET) of crime constitution is a fundamental framework in Chinese criminal law~\cite{liang2017vicissitudes}. It provides a standardized structure to determine criminal liability through four elements: \textbf{Subject, Object, Subjective Aspect, and Objective Aspect}. 

For example, the four-elements of \textit{Robbery} can be briefly summarized as follows:

(1) Subject (the person who commits a criminal act and should bear criminal responsibility): General subjects above the age of criminal responsibility. 

(2) Object (the legal interest harmed by the act): A compound object, combining both property ownership and personal rights of the victim.

(3) Subjective Aspect (the offender's mental state regarding the harmful act): Direct intent with the purpose of unlawfully appropriating another's property.  

(4) Objective Aspect (the external facts of the criminal activity, including key actions and their outcomes): On-the-spot taking of property from an owner, custodian, or possessor through violence, coercion, or other methods. 

For the legal community, FET plays a central role in doctrinal analysis and judicial reasoning. It serves as the legal basis for both legislation and adjudication, ensuring internal consistency and normative rigor in criminal law application~\cite{li2006no_reconstruction, zhang2007normative}. For the legal AI community, FET offers a task-agnostic and interpretable framework for modeling legal reasoning~\cite{deng2023syllogistic, yuan2024can}.

Compared to general reasoning templates (e.g., legal syllogism~\cite{gold2018primer}) or alternative theories such as the Three-Tier System~\cite{zhou2017hierarchical,zhang2010justification}, FET has become the dominant approach in China for assessing criminal liability~\cite{wang2017criminal}. Its clearer and more interpretable decomposition of crimes into objective and subjective elements makes it particularly suitable for structured legal reasoning tasks.

\section{Related Work}
With the rise of open-source base LLMs, lots of legal LLMs have emerged, such as Lawyer LLaMA~\cite{huang2023lawyer}, DiSC-LawLLM~\cite{yue2023disc}, ChatLaw~\cite{cui2024chatlaw}, and TongyiFarui\footnote{\url{https://tongyi.aliyun.com/farui}}. These models are typically adapted from general-purpose LLMs via domain-specific post-training or Retrieval-Augmented Generation (RAG), incorporating legal texts like cases and laws. 

Although these models achieve notable improvements on legal tasks, they still struggle with complex legal reasoning, such as charge disambiguation, legal question answering, statutory interpretation, and structured explanation generation~\cite{hu2025j}. LegalDiscourse shows that LLMs often fail to capture when laws apply and to whom~\cite{spangher2024legaldiscourse}, while LegalBench demonstrates that even state-of-the-art models underperform on diverse reasoning-intensive legal tasks~\cite{guha2023legalbench}.

To further enhance model performance, particularly in tasks requiring complex legal reasoning, some studies draw inspiration from established legal reasoning paradigms. For example, introducing the legal syllogism for legal judgment prediction~\cite{jiang2023legal}; using the IRAC paradigm to guide LLMs in reasoning about compositional rules~\cite{servantez2024chain}. Several works have drawn on the FET in the context of Chinese criminal law. For example, breaking down legal rules into FET-aligned components using automated planning techniques~\cite{yuan2024can}; employing model-generated FETs as minor premises in legal judgment analysis~\cite{deng2023syllogistic}. 

While these methods have demonstrated improved performance on downstream tasks, they generally assume that the LLMs inherently understand the Four-Element Theory, without systematically validating this assumption.

\section{Can LLMs Grasp Legal Theory?}
\label{Can LLM Grasp Legal Theory?}

To examine whether LLMs can understand and apply the Four-Element Theory (FET), we ask them to generate the four elements (FETs) for several representative charges and then analyze the outputs against expert annotations. 

We select GPT-4o as the target LLM, as it achieves state-of-the-art performance on open-source legal benchmarks~\cite{fei2023lawbench,li2024lexeval} and also outperforms others in our pilot study (Appendix~\ref{appendix:pilot study}), indicating a strong capacity to understand and apply legal knowledge. Following prior work~\cite{deng2023syllogistic,cui2024chatlaw,zhou2023boosting}, for each charge, we prompt GPT-4o with corresponding criminal articles (see prompt template in Appendix~\ref{appendix:LLM FET prompt}). 

We invite legal experts who passed the National Judicial Examination to analyze the LLM-generated FETs and identify two main issues: 

(1) Inaccurate elements: LLMs may produce inaccurate FETs. For example, in Figure~\ref{fig:example}, for \textit{misappropriation of public funds}, the LLM-generated Object is ``the management order of public funds and the integrity of officials' conduct'', missing the right to benefit from the use of public funds, which is necessary to identify this charge.

(2) Insufficient interpretive ability: LLMs fail to recognize when statutory language requires deeper interpretation. As shown in Figure \ref{fig:example}, the model simply extracts ``misappropriating public funds for personal use'' to describe the Objective Aspect. However, this phrase is far too general for practice. In judicial interpretations\footnote{National People's Congress Standing Committee. \textit{Interpretation on Article 384, Paragraph 1 of the Criminal Law of the People's Republic of China}, adopted at the 27th Meeting of the Standing Committee of the 9th National People's Congress on April 28, 2002.}, the term ``for personal use'' should be interpreted with three situations: (1) using public funds for oneself, relatives, or other individuals; (2) lending public funds to other entities in one's own name; or (3) using public funds in the name of one's organization for another entity to gain personal benefits. 


\section{Dataset Construction}
The lack of accuracy and interpretation in the generated FETs undermines the reliability of legal reasoning tasks. To address this, we introduce an expert-annotated FET dataset that captures both formal legal definitions and practical interpretive nuances, supporting more trustworthy and adaptable legal AI systems.

To ensure both legal validity and interpretive clarity, we design a hierarchical annotation framework rooted in statutory sources and authoritative interpretive methods. 

\subsection{Hierarchical Legal Interpretation System}

\begin{figure*}[tp]
    \centering
    \includegraphics[width=\textwidth]{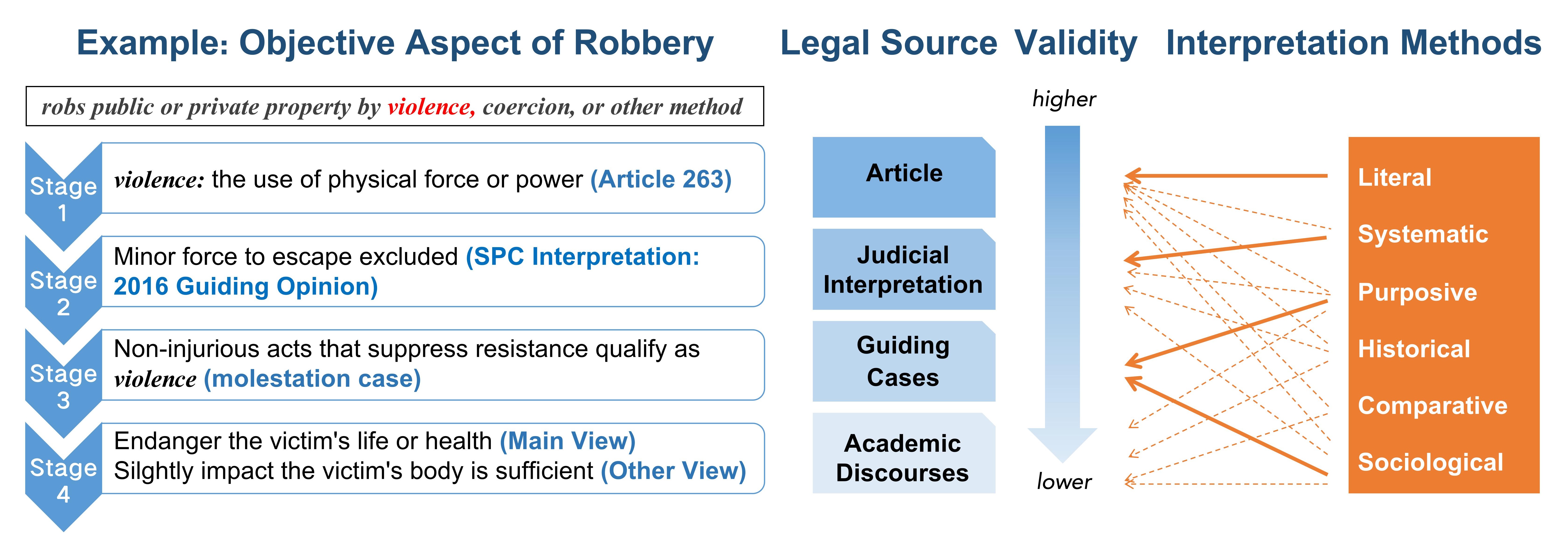}
    \caption{Hierarchical Legal Interpretation System based on legal source validity. The system consists of four annotation rounds, each using different interpretive methods based on different legal sources. Solid arrows indicate the primary method applied; dashed arrows represent supplementary use.}
    \vspace{-2ex}
    \label{fig:main}
    
\end{figure*}
Given a specific charge, we ask five legal experts to annotate the FETs based on relevant legal materials like articles and cases. This annotation process is essentially an act of legal interpretation. \textit{Legal interpretation} refers to the application of various methods to analyze and understand legal texts, to determine their meaning and application in specific legal contexts~\cite{barak2005chapter}. In our task, it involves applying different interpretation methods to the different materials in order to analyze and define the connotation and extension of each of the FETs of a charge. In designing our annotation framework, we address the following two questions:

(1) \textbf{What sources are interpreted.} Legal interpretation draws upon various legal sources with different levels of validity. In legal studies, these sources are categorized based on their legal validity into formal sources (which carry legal force in judgments) and informal sources (which serve as references without legal force)~\cite{pound1925jurisprudence,watson1982legal,pound1932hierarchy}. Articles and judicial interpretations are considered formal sources, whereas case precedents and academic discourses are regarded as informal sources under the Chinese legal system~\cite{zhang2007jurisprudence}. Accordingly, we organize legal sources by their level of validity, with the following order of priority: Article → Judicial Interpretations → Guiding Cases → Academic Discourses.

(2) \textbf{How the law is interpreted.} When interpreting the above sources, different interpretation methods are required. These methods follow a hierarchical order~\cite{sutherland1891statutes,kim2008statutory,eig2014statutory}: Legal interpretation should begin with literal interpretation (interpreting the text based on its plain meaning). If the intended meaning cannot be clearly derived from the article alone, systematic interpretation (considering the article's role within the legal system) and purposive interpretation (considering the legislative intent) should be applied. If ambiguity remains, historical interpretation (based on the legislative history), sociological interpretation (based on the article's social function and consequences), and other interpretation methods may be used to further clarify the legal meaning. The specific definition of legal interpretation methods is in Appendix~\ref{appendix: interpretain}.

We also consider the nature of each source. For example, Guiding Cases, as informal sources, do not define elements literally but instead supplement statutory interpretation through purposive, sociological, and other interpretive methods. The correspondence between interpretation methods and legal sources is illustrated by the orange arrows in Figure~\ref{fig:main}.

\subsection{Annotation Process}

As shown in the left part of Figure~\ref{fig:main}, our annotation process takes charges as input and outputs corresponding FETs, following a Hierarchical Legal Interpretation System to organize legal sources by validity and apply interpretation methods. The annotators are five experts, all of whom have passed the National Judicial Examination and are familiar with FET. The entire annotation process took 7 months and involved four rounds.

\paragraph{Stage One: Literal interpretation using the core article.} 
The interpretation of the FETs begins with the core article of each charge, which carries the highest legal validity, mainly through \textit{literal interpretation}. 

At this stage, annotators analyze the article's subject–predicate–object structure to identify candidate FETs, mapping the subject to the Subject (who commits the crime), the predicate (verb phrase) to the Objective Aspect (the conduct carried out), the object to the Object (the legal interest infringed), and adverbial phrases to the Subjective Aspect (the offender's mental state).

For example, Article 263 of the Chinese Criminal Law, concerning \textit{robbery}, states that ``forcibly seizing public or private property through violence, coercion, or other means'' describes the Objective Aspect. Since the article lacks an explicit subject, a general subject is assumed by default. The adverbs ``violence'' and ``coercion'' indicate an intentional act. This stage spans two months. 

\paragraph{Stage Two: Systematic interpretation using related articles and judicial interpretations.} 
While Stage One relies on the primary article of each charge, literal interpretation often leaves FETs underspecified. Stage Two, therefore, applies \textit{systematic interpretation}, situating underspecified terms within the broader legal framework. By considering the provision's function in the Criminal Law and its links to related articles and judicial interpretations (broadly understood here to include SPC guiding opinions and other interpretative documents), annotators clarify the scope and meaning of the element. 

For example, Article 263 of the Criminal Law does not specify whether ``violence'' must be directed only at persons or may also extend to property, nor does it make clear which borderline conduct should be excluded.  Article 289~\cite{npc2017criminal} provides that in mass ``smashing, looting, and robbing,'' the destruction or seizure of property by ringleaders may be punished as robbery, suggesting that violence may also cover acts against property. Conversely, the SPC's 2016 Guiding Opinion on Robbery~\cite{spc2016robbery} clarifies that if an offender uses only minor force to escape after theft, fraud, or snatching, and no injury above the statutory threshold results, such conduct is not deemed ``violence'' and does not requalify the offense as robbery.

\paragraph{Stage Three: Purposive and sociological interpretation using guiding cases.}
Although the first two stages define the FETs based on articles and judicial interpretations, these sources remain abstract. In legal practice, courts also refer to Guiding Cases, designated by the Supreme People's Court since 2011, to illustrate how legal articles are applied in concrete cases and interpreted in light of social purposes~\cite{chen2024detecting}.  

At this stage, annotators refine FETs by incorporating specific case scenarios from Guiding Cases. Since the number of Guiding Cases is limited, for rarer charges, we also consult model cases from the People's Court Case Database and Gazette cases\footnote{retrieved from PKULaw: \url{https://www.pkulaw.com/case?way=topGuid}}. Annotators mainly apply \textit{purposive and sociological interpretation} to examine how legal elements are concretized in the reasoning process of practical cases, considering both legislative intent and social context.

For example, in defining how ``violence'' in robbery operates in practice, annotators refer to a case involving molestation~\cite{Ma2021Robbery}. The offender bound the victim to commit indecent acts and then took her phone. Under purposive and sociological interpretation, the ongoing molestation maintained the victim's restrained state and thus constituted new violence. Such cases clarify that acts like molestation, though not physically injurious, can suppress resistance and therefore qualify as violence. This stage spans two months.

\paragraph{Stage Four: Diverse interpretations using academic discourses.}
Although Stages One to Three refine both the abstract definitions and concrete scenarios of each element from various legal sources, certain elements still involve unresolved issues that rarely appear in practice and therefore lack clear judicial standards. 

At this stage, annotators consult academic discourses and apply diverse interpretations such as \textit{comparative, purposive, and sociological interpretation}. For elements where disagreement exists, they record both mainstream and minority views, providing concise annotations that explain the underlying legal reasoning.

For example, in defining ``violence'' in robbery, mainstream views in China, the former Soviet Union, North Korea, and Japan require that it endanger the victim's life or health~\cite{Zhang2007}, while others argue that any force sufficient to subdue the victim should qualify~\cite{Yang2010}. The annotations record both the dominant consensus and minority positions. This stage spans one month.

The four stages represent the main interpretive approaches, but they are not mutually exclusive. In practice, annotators often combine methods when clarifying a particular element. As shown in Figure~\ref{fig:main}, the dashed orange arrows mark cross-stage interactions where multiple interpretive methods operate in a complementary way.

\subsection{Data Statistics}
\begin{table}[h]
\centering    
\resizebox{\linewidth}{!}{
\begin{tabular}{l|llll}
\hline
Metric & LLM\textsubscript{Mean} & LLM\textsubscript{Median} & Expert\textsubscript{Mean} & Expert\textsubscript{Median} \\ \hline
Avg. Length & 115.43 & - & 472.53 & - \\
Subject & 23.12 & 27 & 51.64 & 17 \\
Object & 15.86 & 15 & 36.01 & 25 \\
Subjective Aspect & 28.00 & 30 & 42.38 & 21 \\
Objective Aspect & 48.45 & 45 & 342.5 & 230 \\ \hline
\end{tabular}
}
\caption{Comparison of element lengths.}
\label{tab:element_length_comparison}
\end{table}

Our dataset comprises 155 common criminal charges. These charges are selected based on their frequency in over 2.6 million publicly available criminal cases in China: specifically, we include all charges that appear more than 3,000 times, which together account for 91.71\% of all cases (Appendix~\ref{appendix:Detailed Data Distribution for Each Element}), ensuring coverage of the most common real-world judicial scenarios.

To compare the quality of expert-annotated FETs (expert-FETs) and LLM-generated FETs, we selected 105 charges in JUREX-4E that also appear in the widely used LeCaRDv2 dataset~\cite{li2024lecardv2}. LLM-generated FETs were produced using the same setup as before, with a maximum output of 8192 tokens. Table~\ref{tab:element_length_comparison} summarizes the differences in element length, with full length distributions available in Appendix~\ref{appendix:Detailed Data Distribution for Each Element}.

Overall, expert-FETs are significantly longer, with an average total length of 472.53 words compared to 115.43 for LLM-generated ones. The most pronounced gap appears in the Objective Aspect (OA) (mean: 342.5 vs. 48.45), where experts provide detailed factual descriptions, such as action, result, time, and location, often underdeveloped in LLM outputs. While the Subject (SB), Object (OB), and Subjective Aspect (SA) show smaller median differences, notable variation remains, especially in SB (mean: 51.64 vs. 17), which in certain charges involves complex legal interpretations (e.g., ``work'' in copyright infringement) requiring more elaborate legal definitions. 

\section{Human Evaluation}
\label{human eval}
To compare the quality of expert-annotated and LLM-generated FETs, we selected six complicated charges in Chinese judicial practice~\cite{ouyang1999confusing}. Based on prior theoretical framework~\cite{zhang2007normative}, we assess the quality of FETs along four independent dimensions: \textbf{Precision, Completeness, Representativeness, and Standardization}. 

\begin{compactitem}
  \item Precision: Whether each element accurately aligns with its statutory definition, reflecting key terms in the corresponding legal article.
  \item Completeness: Whether each element includes all practically necessary information, ensuring the definition is sufficient to guide legal reasoning. 
  \item Representativeness: Whether the annotations reflect the most typical and practically significant scenarios in judicial practice.
  \item Standardization: Whether the expressions of elements are consistent across different charges, with clear, concise, and unambiguous language that facilitates understanding and minimizes interpretive variance.
\end{compactitem}

Evaluation was conducted by experts from two backgrounds: one group with a purely legal background and another with a combined background in law and AI, all of whom have passed the National Judicial Examination. The experts were selected to balance domain expertise and interdisciplinary perspectives. Scores were averaged across the two groups. Details about the 1-5 scale criteria and annotator background are provided in Appendix~\ref{appendix: human eval guidance}.

\begin{table}[tp]
    \centering
    \small
    \begin{tabular}{l|ccc}
    \hline
    \textbf{Dimension} & \textbf{LLM} & \textbf{Expert} & \textbf{$\delta$} \\
    \hline
    Precision          & 4.12  & 4.69  & + 0.57  \\
    Completeness       & 3.79  & 4.65  & + 0.86  \\
    Representativeness & 3.60  & 4.48  & + 0.88  \\
    Standardization    & 4.33  & 4.56  & + 0.23  \\
    \hline
    \end{tabular}%
    \caption{Performance comparison of four elements across methods. $\delta$ represents the score difference between expert and LLM-generated FETs, with experts outperforming LLMs in all dimensions.}
    \label{tab:Legal Dimension Metric-res}
    \vspace{-2ex}
\end{table}

As shown in Table~\ref{tab:Legal Dimension Metric-res}, expert annotations consistently outperform LLM-generated elements across all four dimensions. The most pronounced deficiencies are observed in Completeness (+0.86) and Representativeness (+0.88). This aligns with our earlier analyses, where expert-generated elements include more factual details and representative descriptions. The gap in Precision (+0.57) suggests a tendency toward vague or legally irrelevant content, while the smaller difference in Standardization (+0.23) shows that LLMs can mimic structural patterns but lack deeper normative consistency. These results demonstrate the importance of expert supervision in providing reliable legal knowledge.

\section{Evaluation on Similar Charge Disambiguation}




To further validate annotation quality,  we introduce the Similar Charge Disambiguation (SCD) task~\cite{yuan2024can,li2024graph}. Given the case fact and a set of similar charges, SCD task requires the model to identify which charge is correct. We evaluate whether similar charges can be effectively distinguished based on their FETs, and whether expert-annotated FETs perform better than LLM-generated FETs.
\begin{table*}[htbp]
\centering
\footnotesize
\begin{tabular}{l|cc|cc|cc|cc}
\hline
\textbf{Model} & \multicolumn{2}{c|}{\textbf{F\&E}} & \multicolumn{2}{c|}{\textbf{E\&MPF}} & \multicolumn{2}{c|}{\textbf{AP\&DD}} & \multicolumn{2}{c}{\textbf{Average}} \\
 & Acc & F1 & Acc & F1 & Acc & F1 & Acc & F1 \\
\hline
GPT-4o & 94.36 & 95.81 & 86.49 & 89.76 & 85.54 & 87.12 & 88.72 & 90.07 \\
GPT-4o+Article & 95.34 & 96.30 & \textbf{92.64} & 93.03 & 88.30 & 89.33 & 92.09 & 92.89 \\
Legal-COT & 94.99 & 96.27 & 90.50 & 90.99 & 87.81 & 88.14 & 89.95 & 90.85 \\
MALR & 94.62 & 95.82 & 86.99 & 86.98 & 87.86 & 88.68 & 89.82 & 90.49 \\
Farui-plus+FET\textsubscript{4o} & 89.09 & 90.27 & 86.32 & 88.00 & 75.90 & 77.67 & 83.77 & 85.31 \\
Farui-plus+FET\textsubscript{Expert} & 89.29 & 90.98 & 86.13 & 87.54 & 76.25 & 78.12 & 83.89 & 85.55 \\
Qwen2.5-72b+FET\textsubscript{4o} & 93.15 & 95.06 & 90.99 & 93.56 & 87.71 & 88.56 & 90.62 & 92.39 \\
Qwen2.5-72b+FET\textsubscript{Expert} & 93.29 & 95.18 & 91.18 & 93.66 & 87.81 & 89.45 & 90.76 & 92.76 \\
GPT-4o+FET\textsubscript{farui} & 94.86 & 96.12 & 91.84 & 92.64 & 89.35 & 89.85 & 92.02 & 92.87 \\
GPT-4o+FET\textsubscript{qwen} & 95.53 & 96.53 & 91.82 & 92.96 & 89.48 & 90.09 & 92.28 & 93.19 \\
GPT-4o+FET\textsubscript{4o+farui+qwen} & 94.97 & 96.24 & 91.84 & 92.73 & 89.69 & 90.12 & 92.17 & 93.03 \\
GPT-4o+FET\textsubscript{4o} & 95.73 & 96.56 & 91.87 & 92.01 & 89.61 & 89.69 & 92.40 & 92.75 \\
GPT-4o+FET\textsubscript{4o + ICL} & 95.74 & 96.36 & 91.84 & 92.01 & 90.48 & 90.63 & 92.69 & 93.00 \\
GPT-4o+FET\textsubscript{Expert} & \textbf{96.06} & \textbf{96.69} & 92.57 & \textbf{93.05} & \textbf{90.53} & \textbf{90.62} & \textbf{93.05} & \textbf{93.45} \\
\hline
\end{tabular}
\caption{\label{tab:charge-level-res}
Performance on the Similar Charge Disambiguation (SCD) task. ``Expert'' refers to our expert-annotated FET, while ``4o'', ``qwen'', and ``farui'' refer to FET generated by different LLMs. Highest results are in bold. 
}
\end{table*}

\subsection{Experiment Settings}


\subsubsection{Dataset and metrics}

We chose the SCD dataset released by~\cite{liu2021everything}. Following previous work~\cite{yuan2024can}, we selected three 2-label classification groups: Fraud \& Extortion (F\&E), Embezzlement \& Misappropriation of Public Funds (E\&MPF), and Abuse of Power \& Dereliction of Duty (AP\&DD). Each charge has over 1.9k cases, with a total of 13,962 cases. The details of the groups are shown in Appendix~\ref{app:scd}. Following previous work~\cite{liu2021everything, yuan2024can}, we use Average Accuracy (Acc) and macro-F1 (F1) as evaluation metrics. 

\subsubsection{Baselines and Methods}

We compared the following baselines: \textbf{GPT-4o~\cite{achiam2023gpt}} and \textbf{GPT-4o+Article}, which explicitly supplies relevant legal articles; \textbf{Legal-COT~\cite{kojima2022large}}, a Chain-of-Thought variant that applies the Four-Element Theory step by step, and \textbf{MALR~\cite{yuan2024can}}, a multi-agent framework that decomposes legal tasks into FET-aligned subtasks. Details are in Appendix~\ref{appendix: SCD baselines}.

Methods: Following Section~\ref{Can LLM Grasp Legal Theory?}, our main model is GPT-4o. We also compared Farui-plus (the latest version of Tongyifarui, representative legal LLM) and Qwen2.5-72B~\cite{bai2023qwen} (representative open-source LLM). To incorporate FET knowledge, each group of similar charges is augmented with four-element descriptions, either generated by LLMs or sourced from JUREX-4E. For example, GPT-4o+FET\textsubscript{LLM} uses LLM-generated FETs, while GPT-4o+FET\textsubscript{Expert} uses expert-annotated ones. The input format is fixed across methods, differing only in the \textit{[four-elements of candidate charges]} (Appendix~\ref{app:scd}).
All experiments are zero-shot, with max\_tokens set to 3,000 (10,000 for Legal-COT and MALR) and a temperature of 0 or 0.0001 in repeated runs.

To further explore generation setups, we also evaluate an ICL variant, GPT-4o+FET\textsubscript{4o + ICL}, where two representative FET exemplars (Theft and Snatching) are provided in the prompt to guide generation.

\subsection{Results}

The SCD results are shown in Table~\ref{tab:charge-level-res}, where we can observe that:

\textbf{Effectiveness of Structured FET Knowledge:} Providing specific structured charge FETs yields the highest accuracy among all legal knowledge integration methods.  
Compared to implicit approaches, such as prompts (GPT-4o+Article, Acc 92.09) or reasoning chains (Legal-COT, Acc 89.95), structured FET knowledge offers more effective support for legal decision-making (e.g., GPT-4o+FET\textsubscript{Expert}, Acc 93.05)

\textbf{Superiority of Expert-Annotated FET:} Expert-annotated FET consistently outperforms LLM-generated FET across three representative LLMs, including FET\textsubscript{farui}, FET\textsubscript{qwen}, FET\textsubscript{4o}, and their combination (FET\textsubscript{4o+farui+qwen}). For example, GPT-4o+FET\textsubscript{Expert} surpasses GPT-4o+FET\textsubscript{4o} by 0.65 in average accuracy and 0.70 in F1-score. 

\textbf{Consistent Gains Across Models:}
Expert-annotated FETs yield consistent performance gains across different SCD models. When applied to Farui-plus, Qwen2.5-72b, and GPT-4o, it improves F1-score by +0.24, +0.37, and +0.70, respectively over their LLM-generated FET baselines.

The ICL-based variant yields consistent improvements over direct prompting (Table~\ref{tab:charge-level-res}), demonstrating the benefit of exemplar guidance, though it still lags behind expert-FETs. We also conducted McNemar's test on paired samples between each LLM-generated FET and the expert-FET, which shows statistically significant improvements ($p<0.05$) across all tasks (see Table~\ref{McNemar} in Appendix~\ref{app:scd}).

\section{Application in Legal Case Retrieval}

In this section, we design a simple expert-guided FET method to apply JUREX-4E to the Legal Case Retrieval (LCR) task, which retrieves relevant cases based on case facts. This task is well-suited for FET because it requires a comprehensive comparison of the four elements across different charges in cases. 


\begin{table*}[htbp]
\centering
\resizebox{\linewidth}{!}{
\begin{tabular}{l|rrr|rrrrr|r}
\toprule
\textbf{Model} & \textbf{NDCG@10} & \textbf{NDCG@20} & \textbf{NDCG@30} & \textbf{R@1} & \textbf{R@5} & \textbf{R@10} & \textbf{R@20} & \textbf{R@30} & \textbf{MRR} \\
\midrule
\textbf{BGE (case\_fact only)} & 0.4737 & 0.5539 & 0.5937 & 0.0793 & 0.2945 & 0.4298 & 0.6500 & 0.7394 & 0.1926 \\
\midrule
\textbf{BGE+FET (Qwen2.5)} & 0.5125 & 0.5858 & 0.6350 & 0.1104 & 0.2870 & 0.4653 & 0.6679 & 0.7836 & 0.2168 \\
\multicolumn{1}{l|}{\quad FET only} & 0.3367 & 0.3971 & 0.4487 & 0.0622 & 0.2006 & 0.3279 & 0.4806 & 0.6037 & 0.1524 \\
\textbf{BGE+FET (Expert, Qwen2.5)} & \textbf{0.5295} & \textbf{0.5979} & \textbf{0.6416} & \textbf{0.1124} & \textbf{0.3122} & 0.4838 & 0.6791 & 0.7824 & \textbf{0.2206} \\
\multicolumn{1}{l|}{\quad FET only} & 0.3354 & 0.4035 & 0.4541 & 0.0849 & 0.1923 & 0.3076 & 0.4839 & 0.6097 & 0.1606 \\
\midrule
\textbf{BGE+FET (GPT-4o)} & 0.5139 & 0.5862 & 0.6291 & 0.0980 & 0.2967 & 0.4769 & 0.6802 & 0.7828 & 0.2140 \\
\multicolumn{1}{l|}{\quad FET only} & 0.3583 & 0.4293 & 0.4798 & 0.0506 & 0.2240 & 0.3644 & 0.5383 & 0.6652 & 0.1453 \\
\textbf{BGE+FET (Expert, GPT-4o)} & 0.5211 & 0.5920 & 0.6379 & 0.1024 & 0.3049 & \textbf{0.4883} & \textbf{0.6885} & \textbf{0.7967} & 0.2155 \\
\multicolumn{1}{l|}{\quad FET only} & 0.3766 & 0.4584 & 0.5111 & 0.0715 & 0.1894 & 0.3709 & 0.5891 & 0.7203 & 0.1624 \\
\bottomrule
\end{tabular}
}
\caption{Performance on the Legal Case Retrieval (LCR) task. The highest results are in bold. ``FET only'' indicates using the four-element descriptions without case facts.}
\label{tab:scr_results}
\end{table*}

\subsection{Dataset and Metrics}

LeCaRDv2~\cite{li2024lecardv2} is the latest version of LeCaRD~\cite{ma2021lecard}, which is widely used in legal tasks~\cite{li2024delta,zhou2023boosting}. It comprises 800 queries and 55,192 candidates extracted from 4.3 million criminal case documents. Following previous work~\cite{qin2024explicitly}, we chose 1390 candidates and used NDCG@10, 20, 30, Recall@1, 5, 10, 20, and MRR as metrics. We also tested different candidate pool settings (see Appendix~\ref{appendix:LCR baselines}). The results are consistent.

\subsection{Baselines and Methods}

We adopt a dense retrieval framework based on BGE-m3~\cite{bge_m3}, a strong embedding model for legal and general-domain texts. Given a query \( q \) and a candidate case \( c \), we compute their vector representations using a shared BGE-m3 encoder. Retrieval is performed by computing cosine similarities between the query and all candidates and selecting the top-k candidates.


To enhance retrieval accuracy, we compare the following three methods:

(1) \textbf{BGE(case\_fact only)}: Standard dense retrieval using only BGE-m3 embeddings of the raw case facts.

(2) \textbf{BGE+FET ($\mathcal{M}_g$)}: We prompt different LLMs $\mathcal{M}_g$ to generate a structured four-element description of each case (case-FET) based solely on its facts, without using external knowledge. These case-FETs are then embedded with BGE-m3 and used to compute similarity. Because the FET abstracts away case-specific details, we combine the original fact-based similarity and the FET-based similarity in a ratio of 7:3.

(3) \textbf{BGE+FET (Expert, $\mathcal{M}_g$)}: An expert-guided FET method that incorporates JUREX-4E to guide case-FET generation. It consists of four steps:
\begin{enumerate}[itemsep=1pt,topsep=0pt,parsep=0pt]
    \item Charge Prediction. A charge prediction model $\mathcal{M}_p$ (Qwen-plus, details see Appendix~\ref{appendix:pilot study}) predicts the set of likely charges $Z=\{z_1,...,z_k\}$ for the query case. 
    \item Expert-FET Matching. Retrieving corresponding charge's four-elements $\{f_z\}_{z\in Z}$ for each predicted charge in JUREX-4E. These provide theoretical guidance for subsequent reasoning.  
    \item Case-FET Generation. Guided by $\{f_z\}$, the LLM $\mathcal{M}_g$ generates case-specific four-elements $fet_c$ for candidate $c$.
    \item Dense retrieval. We embed the generated FETs using BGE-m3 and compute similarity scores as in Method (2), combining both factual and FET-based similarities. 
\end{enumerate}
For the $\mathcal{M}_g$, we chose Qwen2.5-72b and GPT-4o. 
The retrieval framework is implemented with the FlagEmbedding Toolkit\footnote{\url{https://github.com/FlagOpen/FlagEmbedding}} with an RTX 3090. Following prior work~\cite{li2024lecardv2,qin2024explicitly}, we also compare some dense retrieval methods to examine the representativeness of BGE-m3. Results of baselines and prompt templates are available in Appendix~\ref{appendix:LCR baselines}.

\subsection{Results}

The LCR results are shown in Table~\ref{tab:scr_results}, where we can observe that:
\textbf{(1) FET Enhances Retrieval.} Integrating the FET improves retrieval performance across all metrics. For instance, BGE+FET(GPT-4o) improves MRR by 11.11\%, and BGE+FET (Expert, GPT-4o) achieves an even larger gain of 11.89\%, indicating that structured legal theory benefits retrieval quality.
\textbf{(2) Expert Knowledge is Important.} Expert-guided case-FET consistently outperforms LLM-generated variants across both Qwen2.5-72b and GPT-4o backbones. For example, BGE+FET (Expert, GPT-4o) achieves higher Recall@30 (0.7967 vs. 0.7828) and MRR (0.2155 vs. 0.2140). The gap is even larger in the \textit{FET only} setting (e.g., MRR 0.1624 vs. 0.1453 for GPT-4o), demonstrating that expert knowledge captures critical legal reasoning that LLMs may overlook.

We provide a case study in Appendix~\ref{appendix:lcr-case}. It illustrates that the expert-annotated FETs of charges provide practical judgment points and key narratives (e.g., the special subject of \textit{Embezzlement}) that help the LLM focus on essential facts to analyze the case-FET.

\section{Discussion}
\begin{table*}[htbp]
\resizebox{\linewidth}{!}{
\centering
\begin{tabular}{lcccccccc}
\hline
\textbf{Model} & \textbf{F\&E Acc} & \textbf{F\&E F1} & \textbf{E\&MPF Acc} & \textbf{E\&MPF F1} & \textbf{AP\&DD Acc} & \textbf{AP\&DD F1} & \textbf{Avg. Acc} & \textbf{Avg. F1} \\
\hline
GPT-4o+FET\textsubscript{4o} & 95.73 & 96.56 & 91.87 & 92.01 & 89.61 & 89.69 & 92.40 & 92.75 \\
GPT-4o+FET\textsubscript{Collaboration} & 95.99 & 96.56 & 91.51 & 91.61 & 91.02 & 90.95 & 92.84 & 93.04 \\
GPT-4o+FET\textsubscript{Expert} & 96.06 & 96.69 & 92.57 & 93.05 & 90.53 & 90.62 & 93.05 & 93.45 \\
\hline
\end{tabular}
}
\caption{Results of the collaboration study on Similar Charge Disambiguation (SCD).}
\label{tab:collaboration-results}
\end{table*}

To examine whether LLMs can effectively support expert annotation, we designed a preliminary collaboration pipeline that is roughly aligned with our annotation stages.

For each charge, the LLM first retrieved relevant statutory articles (the retrieval of legal sources in stages 1–2). It then extracted the top 10 factual keywords from cases cited by experts (the use of factual cues as in stage 3). Experts validated these keywords for each element, after which the LLM generated refined FET annotations by reasoning over the combined legal texts and factual cues. This pipeline yielded measurable improvements over unguided prompting (Table~\ref{tab:collaboration-results}), illustrating the feasibility of expert-in-the-loop workflows. 

Nevertheless, the study also highlights limitations. High-frequency keywords often lacked discriminative power for similar charges—for example, descriptors such as \textit{``beating''} and \textit{``stabbing''} occurred in both \textit{intentional injury} and \textit{intentional homicide}, reducing their utility. The refined FETs still fell short of the depth and precision of expert annotations. Future improvements could include decomposing FET generation into subtasks aligned with annotation stages and enriching each stage with more authoritative sources, thereby strengthening both coverage and normative awareness.

\section{Conclusion}
This paper presents JUREX-4E, an expert-annotated FET knowledge base built through a structured legal interpretation process and validated on downstream tasks. Grounded in widely accepted interpretive methods, our framework is adaptable across different branches of law and legal traditions, making it applicable beyond Chinese criminal law. Moreover, the structured approach to integrating expert domain knowledge may inspire applications in other areas where expert judgment is critical.

\section{Ethical Considerations}

The datasets used in our evaluation are sourced from publicly available legal datasets, with all defendant information anonymized to ensure privacy. 

\section{Limitations}
Our current knowledge base is limited to 155 charges under Chinese Criminal Law due to the high cost of expert annotation. Future work will explore extending it to other legal domains and jurisdictions. 

Another limitation lies in our current integration of factual and legal information. In the LCR task, although case facts are used to generate FETs, the \textit{FET only} variant excludes the original case facts during retrieval, resulting in performance loss (e.g., MRR 0.1624 vs. 0.2155). This suggests that our current method remains coarse-grained, and more fine-grained fusion strategies, such as multi-agent coordination or retrieval-time integration, deserve future exploration.

\section*{Acknowledgments}
This work is supported in part by Beijing Science and Technology Program (Z231100007423011)

\bibliography{custom}

\appendix

\section{Charge Selection and Detailed Data Distribution}
\label{appendix:Detailed Data Distribution for Each Element}
Charge Selection: To systematically determine charge frequency, we analyzed the CAIL2018 dataset~\cite{xiao2018cail2018}, which contains 2,676,075 criminal cases annotated with 183 criminal law articles and 202 criminal charges and a total of 3,010,000 criminal charges. Apart from few charges that have been merged or changed name, our dataset largely covers all criminal charges from CAIL2018 that have a frequency of over 3,000 (>0.099\%) occurrences. 

\begin{figure*}[]
    \centering
    \begin{minipage}{0.40\linewidth}
        \centering
        \includegraphics[width=\linewidth]{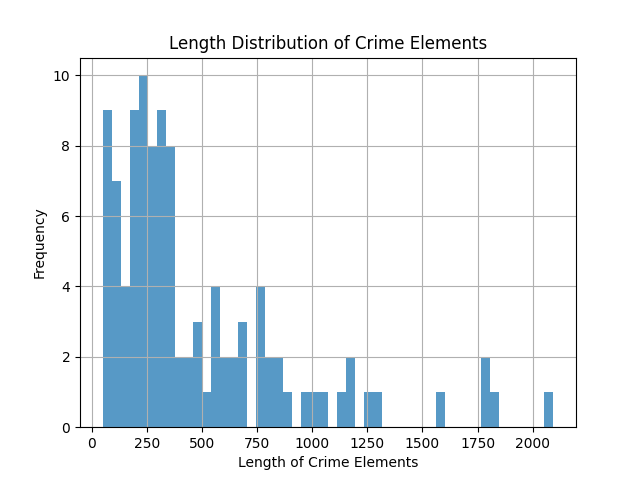}
        \caption{The average length distribution of the total four elements annotated by experts.}
    \end{minipage}%
    \begin{minipage}{0.45\linewidth}
        \centering
        \includegraphics[width=\linewidth]{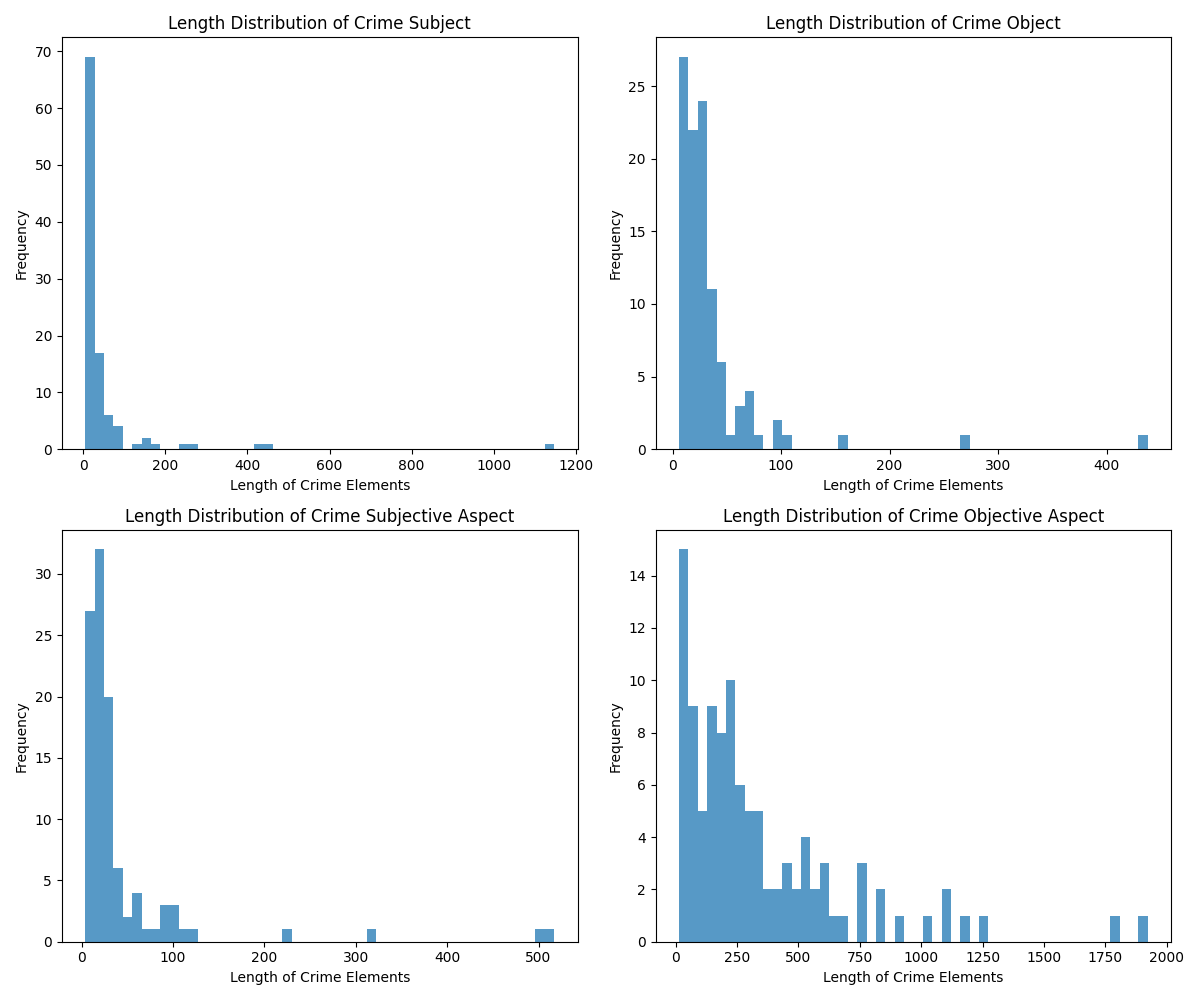}
        \caption{The length distribution of each element annotated by experts.}
    \end{minipage}
    
    \vspace{1cm}
    
    \begin{minipage}{0.45\linewidth}
        \centering
        \includegraphics[width=\linewidth]{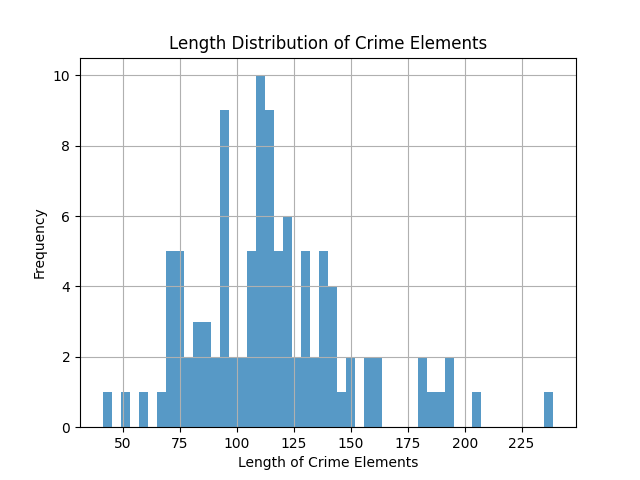}
        \caption{The average length distribution of the total four elements generated by LLM.}
    \end{minipage}%
    \begin{minipage}{0.45\linewidth}
        \centering
        \includegraphics[width=\linewidth]{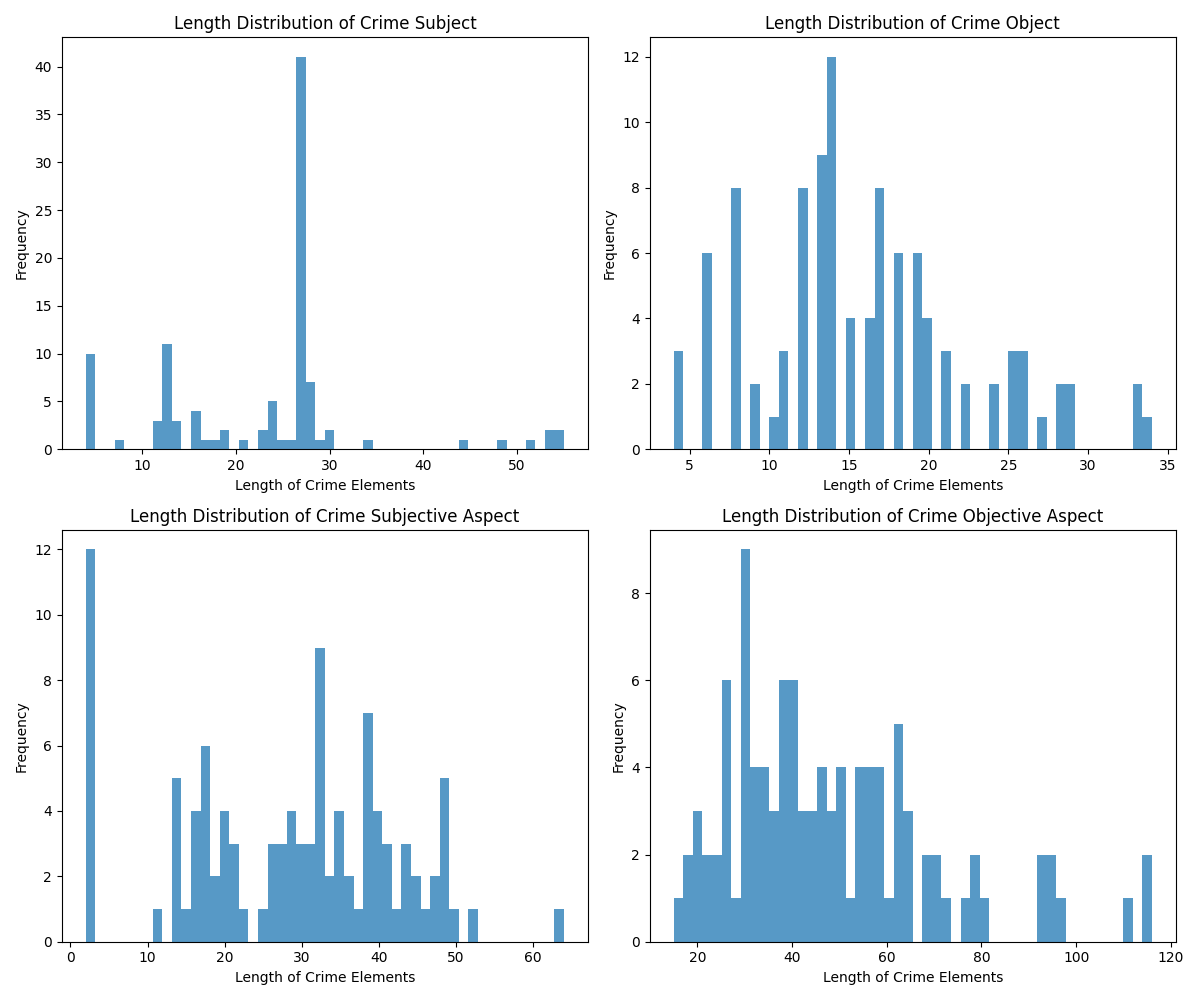}
        \caption{The length distribution of each element generated by LLM.}
        \label{length distribution}
    \end{minipage}
    
\end{figure*}

Length distribution for each element: Table~\ref{length distribution}.

\section{Interpretation Methods}
\label{appendix: interpretain}
1. Literal Interpretation

A strict textual analysis method that adheres to the ordinary meaning of words as understood by a reasonable person at the time of enactment, excluding subjective intent inference

2. Systematic Interpretation

An approach interpreting legal articles through their position within the codified legal hierarchy and logical connections with related norms, maintaining the integrity of the legal system (aligned with Dworkin's "law as integrity" theory).

3. Purposive Interpretation

A method discerning the objective legislative purpose through analysis of statutory structure and functional goals, distinct from subjective legislative intent (following Hart \& Sacks' legal process school).

4. Historical Interpretation

Interpretation based on legislative history materials, including drafts, debates, and official commentaries, while distinguishing original meaning from framers' subjective intentions (as per Brest's original understanding theory).

5. Comparative Interpretation

A methodology referencing functionally comparable legal systems sharing common juridical traditions, employing analogical reasoning while considering local legal culture (developed through Gottfried Wilhelm Leibniz's comparative law framework).

6. Sociological Interpretation

Interpretation evaluating social efficacy through empirical analysis of implementation effects, guided by Pound's sociological jurisprudence principle that ``law must be measured by its achieved results''.

\section{Prompt for LLM-generated FET}
See Table~\ref{tab:llm-fet-prompt}.
\label{appendix:LLM FET prompt}
\begin{table*}[h!]
\resizebox{\linewidth}{!}{%
\begin{tabular}{|p{0.95\textwidth}|}
\hline
You are an expert in criminal law. Based on the given charge, please analyze it according to China's criminal law and output the four elements of the charge in order, including: \\

- \textbf{Object}: The concretization of a certain abstract social interest. For example, the object of charges that infringe on personal rights is the right to life, while the object of property-related charges could be items such as mobile phones or wallets. \\

- \textbf{Objective Aspect}: The objective facts of the criminal act, including the key actions that trigger the charge (e.g., theft, robbery) and the consequences caused by the act (e.g., serious injury, death, property loss). \\

- \textbf{Subject}: Typically, the general subject of the charge, but in some cases, a specific subject is required (e.g., government officials in certain offenses). \\

- \textbf{Subjective Aspect}: The mental state of the perpetrator, such as intent or negligence. \\

\textbf{Relevant Legal Articles:} [] \\

Please synthesize the above information to generate a refined set of four elements that represent the characteristics of the charge. \\

\textbf{Output format:} \{ "Crime": "", "Four-elements of the Crime": \{ "Crime Object": "", "Objective Aspect": "", "Subject": "", "Subjective Aspect": "" \} \} \\

\textbf{Crime:} [] \\
\\
\hline
\end{tabular}%
}
\caption{Prompt template for generating the four elements of a Crime using LLMs}
\label{tab:llm-fet-prompt}
\end{table*}

\section{Details about Pilot Study}
\label{appendix:pilot study}
We selected candidate models from LawBench~\cite{fei2023lawbench} and LexEval~\cite{li2024lexeval}, which contain the broadest and most up-to-date evaluation of legal LLMs. From these, we chose top-performing models such as GPT-4, Qwen-14B-chat, and representative legal-specific LLMs.

For best performance, we used GPT-4o (the latest version of GPT-4 at that time) and Qwen-plus (a stronger commercial variant of Qwen2.5-72B. (Aliyun model-studio official site))

During implementation, we found that most legal LLMs were unavailable. The only stably accessible one was Farui (A leading legal LLM built on Qwen, (Aliyun model-studio official site, Tongyi Farui)), specifically the version ``tongyifarui-890'' from its official API.

To compare GPT-4o, Qwen-plus, and tongyifarui-890, we sampled 300 cases from our legal retrieval dataset and asked models to perform charge prediction, which is the pre-task for generating case-FETs.

(For each case in legal retrieval, the model was required to predict charges, so we can match charges' expert-FETs, and use them to generate case-FET.)

This task involved all criminal charges, including multi-defendant and multi-charge scenarios, and requires models to predict charges from open text without a predefined list, making it a challenging legal task.

The result showed that GPT-4o (59.78\%) > Qwen-plus (58.70\%) >> tongyifarui-890 (21\%). Given Farui's poor performance, we did not include it in subsequent experiments.

We further evaluated GPT-4o and Qwen-plus based on their ability to generate case-FETs. The results showed that GPT-4o outperformed Qwen-plus (MRR 0.2140 vs. 0.2052). Considering both results, we adopted GPT-4o as our primary model in the paper.

Subsequently, in efforts to improve charge prediction for matching charges' expert-FETs, we found that Qwen-plus performed better than GPT-4o when a charge list was provided (58.70\%->80.43\% vs. 59.78\%->71.74\%). Therefore, in this specific setting for charge prediction before retrieval, we used Qwen-plus.

For fair and reproducible presentation of results on specific downstream tasks (SCD and LCR), as mentioned in the main text, we present the results of open source Qwen2.5-72b.

\section{Human Evaluation Guidance}
\label{appendix: human eval guidance}

The annotators included three postgraduate students specializing in criminal law and one master's student in legal science and technology.
The annotators scored independently, without knowledge of each other's results. Before scoring, they were asked to read the descriptions and scoring guidelines (as shown in Table~\ref{tab: four-dimension score details}) for each evaluation dimension. In order to ensure the fairness of the evaluation, they do not know the source of the four elements, and they do not know that the four elements include those generated by LLMs. 

When assigning scores, they were also required to provide brief justifications. For example, for the Completeness dimension: 3 (The description of Objective Aspect is too brief, and does not specify the intent of illegal possession).

\begin{table*}[h]
\centering
\small
\renewcommand{\arraystretch}{1.5} 
\begin{tabularx}{\linewidth}{l|XXXX}
\hline
\multicolumn{1}{c|}{\textbf{Dimension}} & \multicolumn{1}{c}{\textbf{Precision}} & \multicolumn{1}{c}{\textbf{Completeness}} & \multicolumn{1}{c}{\textbf{Representativeness}} & \multicolumn{1}{c}{\textbf{Standardization}} \\ \hline
\textbf{Definition} & Whether there are errors in key elements & Whether the four-elements are complete & Whether key elements and scenarios are emphasized & Whether language and format are clear and standardized \\ \hline
\textbf{Score 1} & Contains numerous obvious errors, severely impeding the judgment of culpability, exculpation, and conviction, leading to significant deviations. & Severe omission of key content, unable to present a complete picture of the crime structure, greatly hindering analysis of criminal behavior. & Completely fails to mention any key elements or scenarios, unable to highlight essential points for crime recognition, offering no assistance in conviction. & Language is extremely chaotic and obscure; format lacks any standardization, greatly hindering comprehension and application. \\
\textbf{Score 2} & Contains multiple noticeable errors, significantly interfering with culpability, exculpation, and conviction judgments, potentially leading to partial errors. & Noticeable omissions in content, failing to comprehensively cover crime elements, affecting thorough analysis of criminal behavior. & Only highlights a minimal and unimportant portion of the key elements, providing weak support for understanding key crime features. & Language is relatively vague and inaccurate, with a casual format that makes content comprehension significantly challenging. \\
\textbf{Score 3} & Contains a few errors, but the overall accuracy in determining culpability, exculpation, and conviction is relatively unaffected, unlikely to lead to judgment errors. & Some key content descriptions are incomplete, but they generally present the framework of the crime structure. & Highlights some relatively important key elements but lacks comprehensiveness and prominence, offering limited assistance in crime identification. & Language is generally clear but may have minor deviations in phrasing or formatting. \\
\textbf{Score 4} & Almost error-free, key elements accurately serve culpability, exculpation, and conviction judgments, ensuring the accuracy of results. & Key elements are mostly complete, with only very slight and non-critical deficiencies that do not hinder a comprehensive analysis of the crime. & Clearly and relatively comprehensively highlights key elements, aiding in accurately identifying crucial aspects of criminal behavior. & Language is clear and accurate, format is relatively standardized, facilitating comprehension and application of relevant content. \\
\textbf{Score 5} & Completely error-free, key elements are precisely defined, achieving highly accurate culpability, exculpation, and conviction judgments without any flaws. & All four elements are complete and detailed, covering every aspect of the crime, perfectly presenting the crime structure. & Precisely and comprehensively highlights all crucial elements, enabling immediate grasp of the core aspects of the crime, significantly aiding conviction. & Language is extremely clear, standardized, and concise; format perfectly meets requirements, with no barriers to understanding, ensuring efficient information delivery. \\ \hline
\end{tabularx}
\caption{The four dimensions of the human evaluation and the specific score description.}
\label{tab: four-dimension score details}
\end{table*}

\section{Details for Similar Charge Disambiguation}
\label{app:scd}

\begin{table}
\renewcommand{\arraystretch}{1.5} 
  \centering
  \small
\begin{tabularx}{\linewidth}{l|X|l}
\hline
\textbf{Charge Sets} & \textbf{Charges} & \textbf{Cases} \\
\hline
F\&E & Fraud \& Extortion & 3536 / 2149 \\
E\&MPF & Embezzlement \& Misappropriation of Public Funds & 2391 / 1998 \\
AP\&DD & Abuse of Power \& Dereliction of Duty & 1950 / 1938 \\
\hline
\end{tabularx}
  \caption{\label{tab:GCI-dataset}
    Distribution of charges in the GCI dataset. Cases denote the number of cases in each category. Following~\cite{liu2021everything}, for a case with both confusable charges, the prediction of any one of the charges is considered correct. 
  }
\end{table}

\begin{table*}[tp]
\centering
\resizebox{\linewidth}{!}{
\begin{tabular}{|l|}
\hline
\textbf{Prompt:} \\
You are a lawyer specializing in criminal law. Based on Chinese criminal law, \\
Please determine which of the following candidate charges the given facts align with. \\
\textcolor{red}{The candidate charges and their corresponding four-elements are as follows:} \\
\textit{\textcolor{red}{[four-elements of Candidate Charges]}}. \\
The four elements represent the core factors for determining the constitution of a criminal charge.\\
\textit{[The basic concepts of the Four-Element Theory]} \\
Please compare the case facts to determine which charge's four elements they align with, thereby identifying the charge. \\
\hline
\end{tabular}}
\caption{Prompt template for adding the Four-Element Theory and specific four-elements of crime in charge disambiguation.}
\label{tab:SCD prompt}
\end{table*}

\label{appendix: SCD baselines}

For LLM baselines, we evaluate both general-purpose and task-specific methods.  

\textbf{GPT-4o} is an optimized version of GPT-4~\cite{achiam2023gpt} that has well performance in specific tasks through domain adaptation. 

To explore the effectiveness of expert-FETs, we further consider other methods that introduced the Four-element Theory into LLMs.

\textbf{GPT-4o\textsubscript{Law}}, which introduces articles related to corresponding charges into the instruction to provide legal context.  

\textbf{Legal-COT} is a variant of COT~\cite{kojima2022large} that guides the LLM to perform step-by-step legal reasoning by incorporating explanations of the Four-element theory into the instruction.

\textbf{MALR} is an up-to-date multi-agent framework designed to enhance complex legal reasoning~\cite{yuan2024can}, enabling LLMs to autonomously decompose legal tasks and extract insights from legal rules. As its full implementation is not publicly available, we use the released code for the auto-planner module and implement the legal insight extraction following the specified steps and prompts, with necessary refinements. Experiments on the paper's reported examples show that our implementation produces task decompositions and outputs largely consistent with the original results.

As shown in Table~\ref{tab:prompt of different method}, different methods differ in their prompts for generating and explaining the Four-Element Theory, but generally follow a similar process. For the SCD output, except for COT and MALR, which require reasoning processes and prediction results, all other methods only require the output of prediction results.

\begin{table*}[htbp]
\resizebox{\linewidth}{!}{
\renewcommand{\arraystretch}{1.5} 
\centering
\begin{tabularx}{\linewidth}{l|XXXXX}
\toprule
\textbf{Method} &
  \textbf{GPT-4o} &
  \textbf{GPT-4o+Article} &
  \textbf{Legal-COT} &
  \textbf{GPT-4o+FET\textsubscript{LLM}} &
\textbf{GPT-4o+FET\textsubscript{Experts}}  \\
\midrule

Pre-task &
  None &
  None &
  None &
  LLM-generated FETs &
  Expert-annotated FETs  \\ 
\midrule

Prompt &
  \multicolumn{5}{p{13cm}}{\raggedright You are a lawyer specializing in criminal law. Based on Chinese criminal law, please determine which of the following candidate charges the given facts align with.} \\ \cline{2-6}

 &
  Candidate charges are as follows: \textit{\textcolor{red}{\#Candidate Charges}} &
  The candidate charges and relevant legal articles are as follows: \textit{\textcolor{red}{\#Candidate Charges + \#Articles}} &
  Please analyze using the four-elements Theory step by step: \textit{\textcolor{gray}{\#details about each step. }}The candidate charges  are as follows: \textit{\textcolor{red}{\#Candidate Charges }} &
  \multicolumn{2}{p{5cm}}{\raggedright The candidate charges and their corresponding four-elements are as follows: \textit{\textcolor{red}{\#four-elements of candidate charges}}. The four elements represent the four core factors of a charge. Compare the case facts to determine which charge's four elements they align with, thereby identifying the charge.} \\  \cline{2-6}

 &
  \multicolumn{5}{p{14cm}}{\raggedright Output format: \textit{\textcolor{orange}{\#Format}}. Note: Only output the charge, no additional information. \\ Case facts: \textit{\textcolor{blue}{\#Case Facts.}}} \\

\bottomrule
\end{tabularx}}
\caption{Prompts of different methods in Similar Charge Disambiguation. \# represents a format input.}
\label{tab:prompt of different method}
\end{table*}

\begin{table*}[ht]
\centering
\begin{tabular}{lccc}
\hline
\textbf{FET-LLM vs FET-Expert} & \textbf{F\&E} & \textbf{E\&MPF} & \textbf{AP\&DD} \\
\hline
FET-qwen vs FET-Expert          & 0.00215 & 0.00070 & 0.00509 \\
FET-farui vs FET-Expert         & 0.00000 & 0.00126 & 0.00516 \\
FET-4o+farui+qwen vs FET-Expert & 0.00000 & 0.00996 & 0.03415 \\
FET-4o vs FET-Expert            & 0.02246 & 0.00156 & 0.02251 \\
\hline
\end{tabular}
\caption{McNemar's test results (p-values) comparing LLM-generated FET and expert-FET. Statistically significant improvements ($p<0.05$) are observed across all tasks.}
\label{McNemar}
\end{table*}

\section{Baselines in Legal Case Retrieval}
\label{appendix:LCR baselines}




\textbf{BERT}~\cite{devlin2018bert} is a language model widely used in retrieval tasks. In this paper, we chose BERT-base-Chinese\footnote{\url{https://huggingface.co/google-bert/bert-base-chinese}}. \textbf{Legal-BERT}\footnote{\url{https://github.com/thunlp/OpenCLaP}}~\cite{chalkidis2020legal} is a variant of BERT that is specifically trained on legal corpora. 
\textbf{Lawformer}~\cite{xiao2021lawformer}is a Chinese legal pre-trained model based on Longformer~\cite{beltagy2020longformer}, which is able to process long texts in the legal domain. 
\textbf{ChatLaw-Text2Vec}\footnote{\url{https://modelscope.cn/models/fengshan/ChatLaw-Text2Vec}}~\cite{cui2023chatlaw} is a Chinese legal LLM trained on 936,727 legal cases for similarity calculation of legal-related texts. 
\textbf{SAILER}~\cite{li2023sailer} is a structure-aware legal case retrieval model utilizing the structural information in legal case documents. 

Baseline results are provided in Table~\ref{tab:full_scr_results}. 

\begin{table*}[htbp]
\centering
\resizebox{\linewidth}{!}{
\begin{tabular}{l|rrr|rrrrr|r}
\toprule
\textbf{Model} & \textbf{NDCG@10} & \textbf{NDCG@20} & \textbf{NDCG@30} & \textbf{R@1} & \textbf{R@5} & \textbf{R@10} & \textbf{R@20} & \textbf{R@30} & \textbf{MRR} \\
\midrule
BERT             & 0.1511           & 0.1794           & 0.1978           & 0.0199          & 0.0753          & 0.1299          & 0.2157          & 0.2579          & 0.1136          \\
Legal-BERT       & 0.1300           & 0.1487           & 0.1649           & 0.0186          & 0.0542          & 0.1309          & 0.1822          & 0.2172          & 0.0573          \\
Lawformer        & 0.2684           & 0.3049           & 0.3560           & 0.0432          & 0.1479          & 0.2330          & 0.3349          & 0.4683          & 0.1096          \\
ChatLaw-Text2Vec & 0.2049           & 0.2328           & 0.2745           & 0.0353          & 0.1306          & 0.1913          & 0.2684          & 0.3751          & 0.1285          \\
SAILER           & 0.3142           & 0.4133           & 0.4745           & 0.0539          & 0.1780          & 0.3442          & 0.5688          & 0.7092          & 0.1427          \\
\textbf{BGE (case\_fact only)} & 0.4737 & 0.5539 & 0.5937 & 0.0793 & 0.2945 & 0.4298 & 0.6500 & 0.7394 & 0.1926 \\
\midrule
\textbf{BGE+FET (Qwen2.5)} & 0.5125 & 0.5858 & 0.6350 & 0.1104 & 0.2870 & 0.4653 & 0.6679 & 0.7836 & 0.2168 \\
\multicolumn{1}{l|}{\quad FET only} & 0.3367 & 0.3971 & 0.4487 & 0.0622 & 0.2006 & 0.3279 & 0.4806 & 0.6037 & 0.1524 \\
\textbf{BGE+FET (Expert, Qwen2.5)} & \textbf{0.5295} & \textbf{0.5979} & \textbf{0.6416} & \textbf{0.1124} & \textbf{0.3122} & 0.4838 & 0.6791 & 0.7824 & \textbf{0.2206} \\
\multicolumn{1}{l|}{\quad FET only} & 0.3354 & 0.4035 & 0.4541 & 0.0849 & 0.1923 & 0.3076 & 0.4839 & 0.6097 & 0.1606 \\
\midrule
\textbf{BGE+FET (GPT-4o)} & 0.5139 & 0.5862 & 0.6291 & 0.0980 & 0.2967 & 0.4769 & 0.6802 & 0.7828 & 0.2140 \\
\multicolumn{1}{l|}{\quad FET only} & 0.3583 & 0.4293 & 0.4798 & 0.0506 & 0.2240 & 0.3644 & 0.5383 & 0.6652 & 0.1453 \\
\textbf{BGE+FET (Expert, GPT-4o)} & 0.5211 & 0.5920 & 0.6379 & 0.1024 & 0.3049 & \textbf{0.4883} & \textbf{0.6885} & \textbf{0.7967} & 0.2155 \\
\multicolumn{1}{l|}{\quad FET only} & 0.3766 & 0.4584 & 0.5111 & 0.0715 & 0.1894 & 0.3709 & 0.5891 & 0.7203 & 0.1624 \\
\bottomrule
\end{tabular}
}
\caption{Performance on the Legal Charge Retrieval (LCR) task with baselines. Highest results are in bold. ``FET only'' indicates using the four-element descriptions without case facts.}
\label{tab:full_scr_results}
\end{table*}

To support reproducibility, we provide the full prompt templates used in our pipeline. Table~\ref{tab:prompt-charge} shows the prompt for charge prediction, and Table~\ref{tab:prompt-fet} presents the prompt used for generating four-element annotations in both BGE+FET(LLM) and BGE+FET(Expert, LLM).

\begin{table*}[htbp]
\centering
\begin{tabular}{|p{0.96\textwidth}|}
\hline
\textbf{Prompt 1: Charge Prediction} \\ \hline
You are a legal expert specializing in criminal law. Based on the provided list of charges, determine which charges are applicable to the given case facts. Please note that you should only output the charge names, without any additional information. The charges must be selected from the provided list and should be separated by commas. \\

\vspace{0.3em}
\textbf{[Crime List]} \\
\textbf{[Case Facts]} \\
\\
\hline
\end{tabular}
\caption{Prompt used for charge prediction.}
\label{tab:prompt-charge}
\end{table*}

\begin{table*}[ht]
\centering
\begin{tabular}{|p{0.96\textwidth}|}
\hline
\textbf{Prompt 2: BGE+FET(LLM) and BGE+FET(Expert, LLM).} \\ \hline
You are a legal expert specializing in criminal law. Based on Chinese criminal law knowledge, analyze the following case facts and provide the following information in sequence:

1. The four elements of the crime:\\
\ \ - Criminal Object: The tangible or intangible interests being infringed upon (e.g., personal rights such as life, or property rights such as money, vehicles). \\
\ \ - Objective Aspect: The objective facts of the criminal activity, including key actions (e.g., theft, robbery) and consequences (e.g., injury, death, loss). \\
\ \ - Criminal Subject: Typically general subjects; special subjects in certain crimes (e.g., government officials). \\
\ \ - Subjective Aspect: Whether the act was intentional or negligent. \\

2. Charge: Only output the specific crime name(s). \\
3. Relevant Legal Articles: Only output the article number(s) of the relevant laws. \\

Output format: JSON. For each crime involved in the case, provide a separate dictionary entry.

[Output Sample] \\
\{ \\
\ \ "Crime 1": \{ \\
\ \ \ \ "Four elements": \{ \\
\ \ \ \ \ \ "Criminal Object": "Personal rights: the victim Wang's right to life; Property rights: vehicle.", \\
\ \ \ \ \ \ "Objective Aspect": "The defendant Wu drove under the influence and collided with the victim Wang, causing Wang's immediate death and vehicle damage.", \\
\ \ \ \ \ \ "Criminal Subject": "Defendant Wu, the driver.", \\
\ \ \ \ \ \ "Subjective Aspect": "Negligence" \\
\ \ \ \ \}, \\
\ \ \ \ "Charge": "Traffic Accident Crime", \\
\ \ \ \ "Relevant Legal Article": "Article 133" \\
\ \ \}, \\
\ \ "Crime 2": \{ \\
\ \ \ \ "Four elements": \{ \\
\ \ \ \ \ \ "Criminal Object": "Social management order: infringement on the state's document management system; Property rights: forged documents and related items.", \\
\ \ \ \ \ \ "Objective Aspect": "Defendant 1 purchased equipment and materials to forge documents. Defendant 2 delivered the forged documents. Defendant 3 facilitated transactions via the internet, handling payments and document transfers.", \\
\ \ \ \ \ \ "Criminal Subject": "Multiple defendants, all individuals with full criminal responsibility.", \\
\ \ \ \ \ \ "Subjective Aspect": "Intentional" \\
\ \ \ \ \}, \\
\ \ \ \ "Charge": "Forgery, Alteration, or Sale of Official Documents, Certificates, and Seals of State Organs", \\
\ \ \ \ "Relevant Legal Article": "Article 280, Paragraph 1" \\
\ \ \} \\
\}
\} \\
\\
\hline
\end{tabular}
\caption{Prompt for generating four-element annotations used in FET\textsubscript{LLM} and FET\textsubscript{Expert\_Guided}.}
\label{tab:prompt-fet}
\end{table*}

\section{SCR results on the full LeCaRDv2 Dataset}

As presented in Table~\ref{tab:full set SCR results}, we selected several representative methods based on sparse retrieval and dense retrieval for experiments on the full LeCaRDv2 dataset. All language models were not fine-tuned. The expert-guided FET method achieved the best performance among all language models, attaining top results in both R@500 and R@1000. The results indicate that the conclusions drawn from the full dataset are consistent with those from the subset, and the expert-guided method demonstrates strong performance.

\begin{table*}[htbp]
\centering
\begin{tabular}{l|llll}
\hline
Model & \textbf{R@100} & \textbf{R@200} & \textbf{R@500} & \textbf{R@1000} \\ \hline
Legal-BERT & 0.1116 & 0.1493 & 0.2174 & 0.2819 \\
Lawformer & 0.2432 & 0.304 & 0.4054 & 0.4833 \\
ChatLaw-Text2Vec & 0.1045 & 0.1628 & 0.2791 & 0.3999 \\
SAILER & 0.2834 & 0.4033 & 0.6104 & 0.7568 \\ \hline
BGE & 0.4085 & 0.5246 & 0.6855 & 0.7912 \\
BGE+FET(GPT-4o) & 0.4167 & 0.5388 & 0.7006 & 0.7925 \\
BGE+FET(Expert, GPT-4o) & \textbf{0.4201} & 	\textbf{0.5396} & \textbf{0.7010} & \textbf{0.7927} \\ \hline
\end{tabular}
  \caption{SCR results on the full set of LeCaRDv2. Bold fonts indicate leading results in each setting. The expert-guided FET method achieved the best performance among all language models and attained the top results in both R@500 and R@1000. }
  \label{tab:full set SCR results}
\end{table*}

\section{A Case Study of LCR}
\label{appendix:lcr-case}
Table~\ref{tab:LCR case} presents a case study on the Crime of Embezzlement. By comparing the expert-FETs for the charges in JUREX-4E, the case-FETs generated directly by the LLM, and those generated by the LLM with expert-FETs of charge as guidance, we can observe that:

1) Incorporating expert fine-grained annotations enables the model to better grasp the elements of a crime, thereby providing more precise element comparison. For example, LLMs can identify the ``integrity of official duties'', and the subjective aspect ``Intentional'' can be interpreted as ``having the purpose of illegally possessing public or private property'', highlighting the characteristics of ``official duties''. Capturing the core information of the case is crucial for matching cases with similar facts.

2) LLMs can conduct case-tailored specific analysis based on the constitutive elements of a crime. Blue parts show the LLMs can better analyze the defendant's workplace and the actions taken in the case, which reflects the significance of specific and accurate legal knowledge.

\begin{table*}[]
\renewcommand{\arraystretch}{1.5} 
\begin{tabularx}{\linewidth}{l|X|X|X}
\hline
\textbf{Document} & \multicolumn{3}{p{13cm}}{\textit{[Head of document]}...In April 201X, Company A appointed B as the Sales Manager and Deputy Manager of the Catering Department, responsible for collecting outstanding debts. From December 201X to April 201Y, B embezzled X yuan from six units' payments for personal investment and consumption. In May 201Y, knowing that the company required him to reconcile accounts with the finance department, B fled after changing his phone number. It was found that after the case broke out, B's father C had returned the X yuan to Company A.} \\ \hline
\textbf{Charge} & \multicolumn{3}{p{13cm}}{Crime of Embezzlement} \\ \hline
 & \textbf{Expert-FETs} & \textbf{LLM-generated} & \textbf{Expert-guided} \\ \hline
\textbf{FETs} &
  \begin{tabular}[t]{@{}p{4cm}@{}}
  \textbf{Object:} The ownership of property and \textcolor{red}{integrity of duties} in companies, enterprises, or other organizations. The objects include movable and immovable, tangible and intangible property of the perpetrator's organization.\\
  \textbf{Objective Aspect:} Using positional convenience to illegally appropriate organizational property in a significant amount.\\
  \textbf{Subject:} \textcolor{red}{Special subject}, namely personnel of companies, enterprises, or other organizations.\\
  \textbf{Subjective Aspect:}Intentional, \textcolor{red}{with the purpose of illegally possessing property}.
  \end{tabular} &
  \begin{tabular}[t]{@{}p{4cm}@{}}
  \textbf{Object:} The ownership of property of companies, enterprises, or other organizations.\\
  \textbf{Objective Aspect:} Using the convenience of one's position to illegally appropriate the property of the organization, in a significant amount.\\
  \textbf{Subject:} Personnel of companies, enterprises, or other organizations. The defendant is B.\\
  \textbf{Subjective Aspect:} Intentional.
  \end{tabular} &
  \begin{tabular}[t]{@{}p{4cm}@{}}
  \textbf{Object:} The ownership of property of companies, enterprises, or other organizations and \textcolor{red}{the integrity of official duties}. In this case, crime objects are the properties of the company, enterprise, or other organization where the perpetrator is employed.\\
  \textbf{Objective Aspect:} Using one's position for personal gain by illegally appropriating the property of the organization, where the amount involved is significant. \textcolor{blue}{Specifically, B used his position to embezzle X yuan from the payments of six units for personal investment and consumption.}\\
  \textbf{Subject:} \textcolor{red}{Special subject}, namely personnel of companies, enterprises, or other organizations. \textcolor{blue}{Defendant B is the Sales Manager and Deputy Manager of the Catering Department of the company.}\\
  \textbf{Subjective Aspect:} Intentional, \textcolor{red}{with the purpose of illegally possessing public or private property}.
  \end{tabular} \\ \hline
\end{tabularx}
\caption{Results of the case-FETs using different methods for LCR. 
\textbf{Expert-FETs} refer to expert annotations of the predicted charges. 
\textbf{LLM-generated} and \textbf{Expert-guided} indicate whether the Expert-FETs of predicted charges are used to guide the LLM in generating the case-FETs. 
\textcolor{red}{Red} parts indicate knowledge from experts, while \textcolor{blue}{blue} parts represent the LLM's internal knowledge. 
By incorporating JUREX-4E, the model better emphasizes conviction- and sentencing-related information and provides more detailed descriptions of key case facts.}
\label{tab:LCR case}

\end{table*}

\end{document}